\documentclass[journal]{IEEEtran} 
\IEEEoverridecommandlockouts
\usepackage{cite}
\usepackage{amsmath,amssymb,amsfonts}
\usepackage{algorithmic}
\usepackage{graphicx}
\usepackage{textcomp}
\usepackage{xcolor}
\usepackage{booktabs}
\usepackage{subfig}
\usepackage{url}

\def\BibTeX{{\rm B\kern-.05em{\sc i\kern-.025em b}\kern-.08em
    T\kern-.1667em\lower.7ex\hbox{E}\kern-.125emX}}
\begin{document}

\title{Semi-supervised Chinese Poem-to-Painting Generation via Cycle-consistent Adversarial Networks}

\author{Zhengyang Lu, Tianhao Guo, Feng Wang
    \\
    \IEEEauthorblockA{\textit{Jiangnan University, Wuxi, China}}\\
}

\maketitle

\begin{abstract}
Classical Chinese poetry and painting represent the epitome of artistic expression, but the abstract and symbolic nature of their relationship poses a significant challenge for computational translation. Most existing methods rely on large-scale paired datasets, which are scarce in this domain. In this work, we propose a semi-supervised approach using cycle-consistent adversarial networks to leverage the limited paired data and large unpaired corpus of poems and paintings. The key insight is to learn bidirectional mappings that enforce semantic alignment between the visual and textual modalities. We introduce novel evaluation metrics to assess the quality, diversity, and consistency of the generated poems and paintings. Extensive experiments are conducted on a new Chinese Painting Description Dataset (CPDD). The proposed model outperforms previous methods, showing promise in capturing the symbolic essence of artistic expression. Codes are available online \url{https://github.com/Mnster00/poemtopainting}.
\end{abstract}


\section{Introduction}

Classical Chinese poetry and painting represent an important part of the world's cultural heritage, offering a window into ancient Chinese aesthetics, philosophy, and values. The interplay between these two art forms has fascinated artists and scholars for centuries, with paintings often inspired by and embodying the imagery and sentiments expressed in poems. Generating pictorial illustrations of classical Chinese poetry, therefore, presents an intriguing challenge for computational creativity.

The primary characteristic of classical Chinese poetry is highly symbolic and abstract language, where the poet often seeks to evoke a mood or convey a profound meaning through succinct and vivid imagery. This is in contrast to the more descriptive style of most existing datasets used for text-to-image synthesis, such as MSCOCO \cite{lin2014microsoft} and CUB \cite{wah2011caltech}. The artistic style and visual elements in classical Chinese paintings are also quite distinct from photorealistic images. As such, directly applying models trained on natural images and descriptions to the poem-to-painting domain yields unsatisfactory results, as they fail to capture the implicit alignment of abstract concepts.

Another significant challenge is the lack of large-scale paired training data. While millions of poems and paintings from ancient China are preserved, the number of poems with explicitly corresponding paintings is quite limited. Most existing cross-modal translation approaches rely on supervised learning from paired data, which is infeasible in this low-resource setting. There is a need for techniques that can effectively learn from both the small number of paired examples and the larger unpaired corpus.

\begin{figure}
	\centering
	\includegraphics[width=\linewidth]{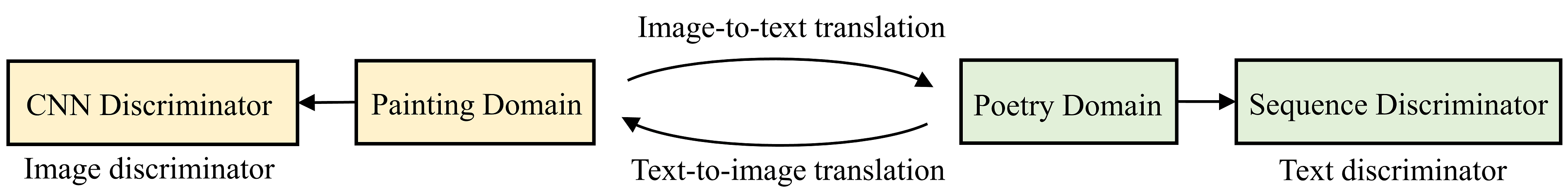}
	\caption{The framework of the proposed semi-supervised framework.}
	\label{fig:cycoverall}
\end{figure}

To address these challenges, we propose a semi-supervised framework for classical Chinese poem-to-painting translation using cycle-consistent adversarial networks. Our approach is inspired by unsupervised image-to-image translation \cite{zhu2017unpaired}, which learns bidirectional mappings to enforce cycle consistency. As shown in Fig.\ref{fig:cycoverall}, we extend this idea to the cross-modal setting by introducing poem and painting encoders that map into a shared semantic space, and corresponding generators that decode from this space. The encoders and generators are trained with both adversarial and cycle consistency losses, ensuring that the generated paintings and poems are realistic and faithful in reconstruction. The use of a shared latent space encourages the network to learn a semantic alignment between the visual and textual modalities.

To our knowledge, ours is the first work to explore semi-supervised poem-to-painting translation with explicit cycle consistency. The main contributions are summarized as follows:
\begin{itemize}
	\item We propose a semi-supervised framework for Chinese poem-to-painting translation using cycle-consistent adversarial networks, which enables the joint learning from both paired and unpaired data.
	\item We introduce several novel evaluation metrics to assess the quality, diversity, and semantic consistency of the generated poems and paintings, drawing insights from human artistic evaluation.
	\item We contribute a new Chinese Painting Description Dataset, providing a valuable resource for research on artistic cross-modal translation.
	\item Extensive experiments on the proposed dataset demonstrate the superiority of our approach against previous methods in generating high-quality, diverse, and semantically meaningful poem-painting pairs.
\end{itemize}

\section{Related works}

Significant progress has been made in text-to-image and image-to-text translation in recent years, driven by advances in deep learning and generative models. 
In this section, we review most relevant works to our proposed approach for poetry and painting generation.

\subsection{Image-to-text translation}

Image-to-text translation aims to generate natural language descriptions from visual input. Early approaches relied on template-based methods that filled in handcrafted templates with detected visual concepts \cite{kulkarni2013babytalk,li2011composing,dan2024multiple}. More recently, deep learning models have achieved significant progress by learning the mapping between images and text in an end-to-end manner \cite{vinyals2015show,xu2015show,lu2022single}.

Encoder-decoder architectures have emerged as a popular choice for image-to-text translation, where convolutional neural networks (CNNs) are employed to encode the image into dense feature vectors, and recurrent neural networks (RNNs) are used to decode these features into word sequences \cite{vinyals2015show}. The incorporation of attention mechanisms has further enhanced the performance of these models by enabling them to selectively focus on relevant image regions when generating each word \cite{xu2015show}. Hierarchical approaches have also been proposed to decompose the generation process into multiple stages, such as first predicting a semantic layout and then filling in the details \cite{krause2017hierarchical,tan2019expressing,dan2024evaluation}. Additionally, adversarial training techniques have been explored to improve the naturalness and diversity of the generated captions \cite{dai2017towards}.

Beyond generating purely descriptive captions, some works have explored more artistic and stylized text generation from images, such as composing poetry \cite{liu2018beyond,zhang2014chinese} or generating stylized captions \cite{mathews2018semstyle,chen2018factual}. However, these approaches often rely on paired image-text datasets for fully-supervised training, which limits their applicability to niche domains like classical Chinese art where such paired data is scarce. To address this limitation, unsupervised or weakly-supervised approaches that can learn from unpaired image and text data have gained attention in recent years \cite{feng2019unsupervised}. These methods typically employ techniques such as cycle consistency \cite{zhu2017unpaired} or adversarial alignment \cite{huang2018multimodal} to bridge the gap between the visual and textual domains, enabling the generation of stylized or artistic descriptions even in the absence of paired training data.

\subsection{Text-to-image generation}

Text-to-image generation aims to synthesize images from natural language descriptions. Generative adversarial networks (GANs) \cite{goodfellow2014generative, lu2022pyramid, lu2024self} revolutionized this field, enabling the generation of realistic images conditioned on textual input.
GANs formulate the problem as a minimax game between a generator network that aims to synthesize realistic images from input texts, and a discriminator network that tries to distinguish between real and generated images. The generator is trained to fool the discriminator, while the discriminator is trained to improve its classification accuracy. This adversarial training paradigm has led to significant improvements in the quality and diversity of generated images.

Reed et al. \cite{reed2016generative} first proposed an end-to-end GANs architecture for text-to-image synthesis, which learned a direct mapping from textual descriptions to images. However, the generated images often lacked fine-grained details and consistency with the input texts. Subsequent works have focused on improving the visual quality and semantic alignment of the generated images through various techniques, such as attention mechanisms \cite{xu2018attngan}, multi-stage refinement \cite{zhang2017stackgan}, and hierarchical generation \cite{zhang2018stackgan++}. 
Qiao et al. \cite{qiao2019mirrorgan} proposed a mirror structure that reconstructed the input text from the generated image, encouraging semantic consistency.
To overcome the paired data limitation, some recent works have explored unsupervised or semi-supervised approaches. Gu et al. \cite{gu2019unpaired} proposed a cycle-consistent adversarial network that learns bidirectional mappings between the text and image domains, enabling text-to-image synthesis without paired data. Huang et al. \cite{huang2019hierarchical} introduced a semi-supervised approach that leverages both paired and unpaired data using a hierarchical alignment strategy.

Beyond GANs, several other advanced approaches have recently emerged for text-to-image synthesis. Diffusion models, such as Denoising Diffusion Probabilistic Models (DDPM) \cite{ho2020denoising}, have shown impressive results in generating high-quality images from text. DDPM learns to iteratively denoise a Gaussian noise input conditioned on the text embedding, generating realistic images through a gradual refinement process.
Autoregressive models, like DALL-E \cite{ramesh2021zero}, have also demonstrated remarkable performance in text-to-image generation. DALL-E uses a transformer architecture to autoregressively predict image tokens conditioned on the input text, enabling the generation of diverse and semantically consistent images.
CogView \cite{ding2021cogview} is another powerful text-to-image model that combines the strengths of autoregressive transformers and variational autoencoders. It learns a joint distribution over text and image tokens, allowing for controllable and high-quality image generation guided by textual descriptions.

For artistic text-to-image synthesis, such as painting generation from poetry, preserving the artistic style and abstract content is essential. Zhu et al. \cite{zhu2019dm} introduced a memory-based model that selectively uses learned artistic strokes and textures to compose paintings that match the poetic descriptions.
Xue et al. \cite{xue2021end} proposed Sketch-And-Paint GAN (SAPGAN), the first end-to-end model for generating Chinese landscape paintings without conditional input, using a two-stage GAN architecture to generate edge maps and translate them into paintings. Fu et al. \cite{fu2021multi} introduced a Flower-Generative Adversarial Network framework to generate multi-style Chinese flower paintings, using attention-guided generators and discriminators, and a novel Multi-Scale Structural Similarity loss to preserve image structure and reduce artifacts.

\begin{figure}
	\centering
	\includegraphics[width=\linewidth]{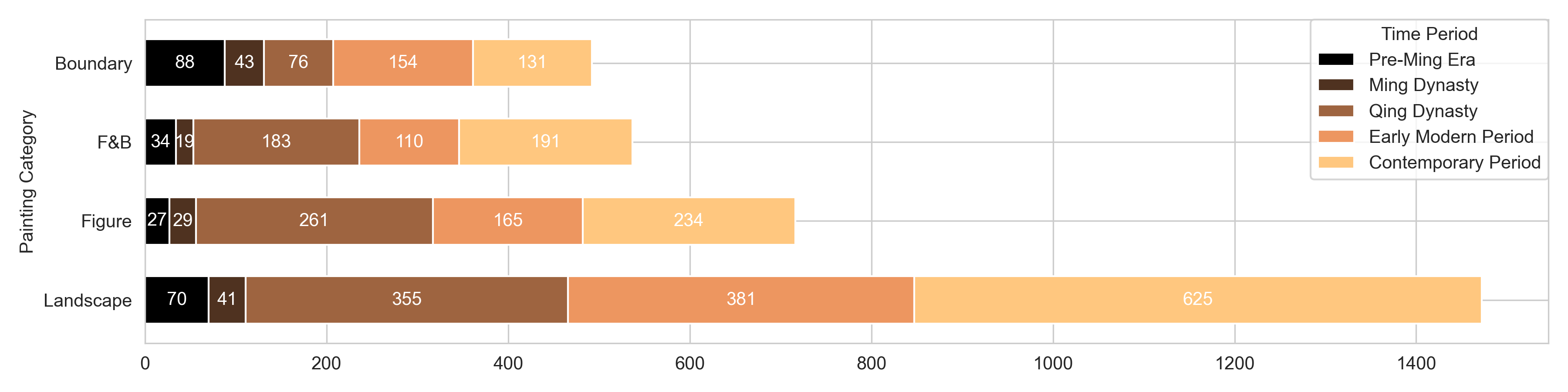}
	\caption{Painting Categories Across Historical Periods}
	\label{fig:paintingstat}
\end{figure}

\begin{figure*}
	\centering
	\includegraphics[width=\linewidth]{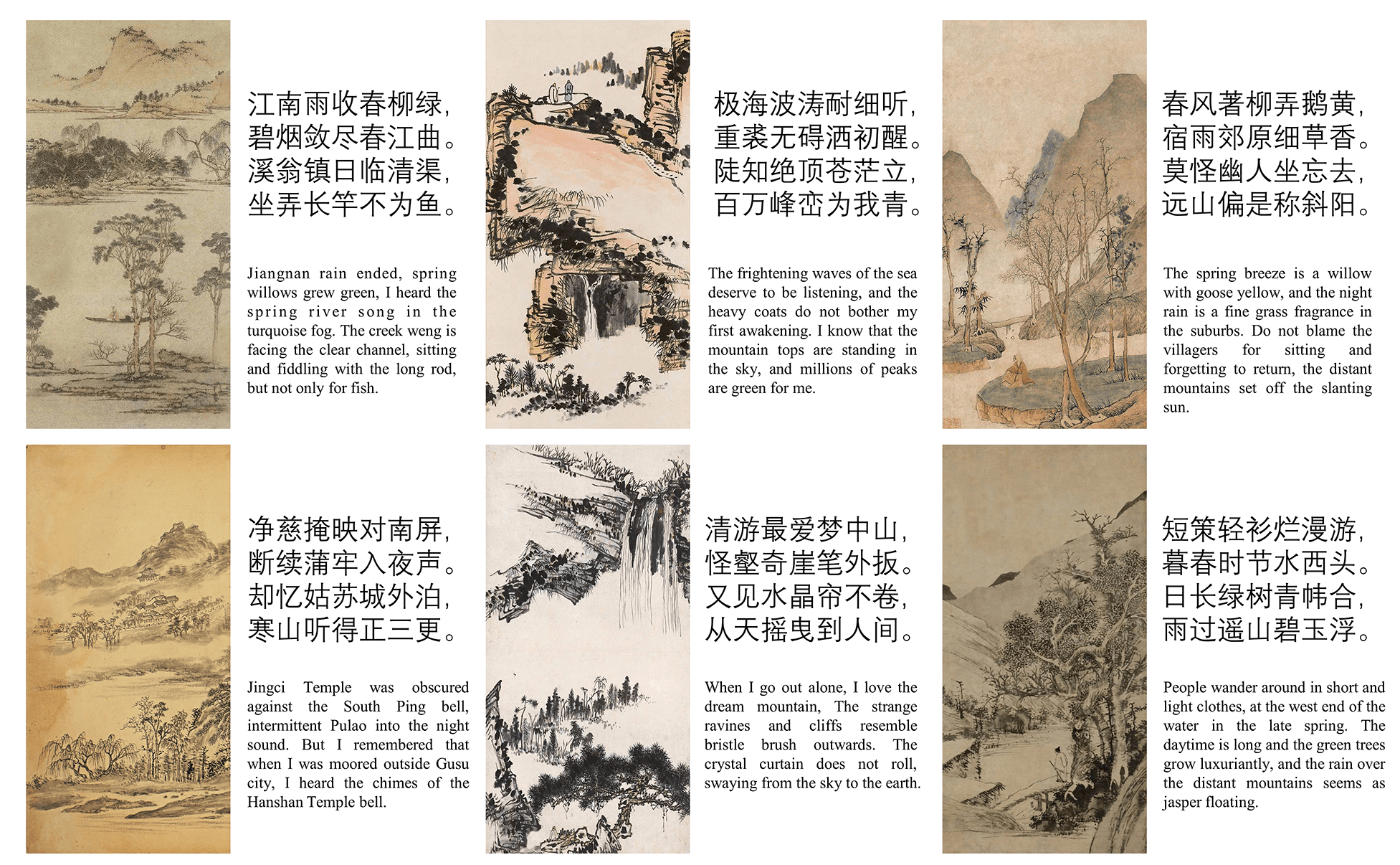}
	\caption{Examples of pairwise poems and paintings from CPDD dataset. English descriptions are literal translations of the original Chinese poems.}
	\label{fig:paintingdemo}
\end{figure*}

\section{Chinese Painting Description Dataset}

High-quality Chinese painting samples are scarce due to the rare large-scale collection in the Chinese traditional art field. Besides the independent painting dataset, Chinese painting and poetry pairs are extremely rare specimens, which only co-exist on paintings with poems inscribed.

To complement dataset absence in the traditional art field, we create Chinese Painting Description Dataset (CPDD), including 3,217 Chinese poems and corresponding paintings whose size are resized to 512$\times$1024. 
According to classic art theory, Chinese paintings are classified into four categories: figure, flower and bird, landscape and boundary paintings.
Hence, the proposed dataset comprises 716 figure paintings, 537 flower and bird paintings, 1,482 landscape paintings, and 492 boundary paintings across various dynasties, as shown in Figure \ref{fig:paintingstat}. 
Within dataset, we have limited the number of modern Chinese paintings due to inconsistent quality and unrecognized styles. 
Therefore, the proposed dataset maintains a balanced representation of Chinese paintings across different historical periods.
The CPDD dataset will be released under a Creative Commons Attribution 4.0 International (CC BY 4.0) license. The poems and paintings in the dataset are in the public domain due to their age, but our curated pairings and annotations are made available under the CC BY 4.0 license.

For Chinese arts, namely Chinese painting and ancient poetry, most expression forms are implicit, as shown in the Figure \ref{fig:paintingdemo}.
In artistic words, all words of scenery are words of feeling. 
Therefore, the proposed dataset is dedicated to improving machine perception of abstract art, aiming to enable deciphering of high-level semantic information and human emotions.

In addition to the paired data from CPDD, we utilized unpaired images from the Wikiart Chinese Painting Collection (WCPC) \cite{phillips2011wiki} and unpaired poems from the Quantangshi Corpus (QC) \cite{broadwell2019reading}. The WCPC contains 25,591 high-resolution images of traditional Chinese paintings spanning various dynasties and genres. The QC consists of 42,863 classical Chinese poems from the Tang Dynasty. In Table \ref{tab:data_sources}, we summarized all training data sources, including sample counts and percentage contributions.

\begin{table}[htbp]
	\centering
	\caption{Summary of Training Data Sources}
	\label{tab:data_sources}
	\begin{tabular}{lrrr}
		\hline
		\textbf{Dataset} & \textbf{Sample Count} & \textbf{Percentage} & \textbf{Type} \\
		\hline
		CPDD (Images+Poems) & 3,217 & 4.49\% & Paired \\
		WCPC (Images) & 25,591 & 35.71\% & Unpaired \\
		QC (Poems) & 42,863 & 59.81\% & Unpaired \\
		\hline
		\textbf{Total} & 71,671 & 100.00\% & - \\
		\hline
	\end{tabular}
\end{table}

\section{Proposed method}

In this section, we describe the proposed poem-to-painting model with cycle-consistent adversarial networks. 
Note that pairs of images and poems are obtained from a manually collected dataset, called Chinese Painting Description Dataset (CPDD), which represents a medium-scale of the training data.
Moreover, the majority of training samples, that is, images and poems, are from separate datasets.
Our goal is to learn to compose high-quality and diverse poems from a single image in a semi-supervised manner.

\begin{figure*}
	\centering
	\includegraphics[width=1\linewidth]{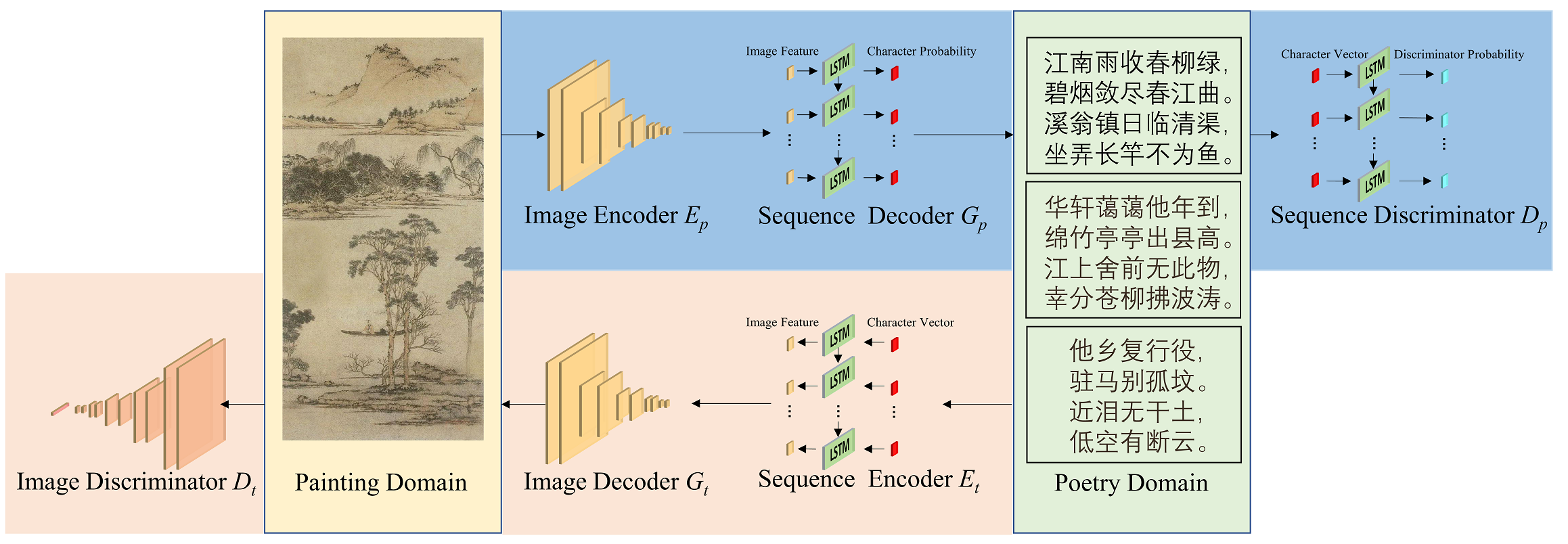}
	\caption{The framework of the proposed cycle-consistent adversarial network for Chinese poem-to-painting translation. It consists of poem and painting encoders ($E_p$, $E_t$) that map into a shared latent space, corresponding generators ($G_p$, $G_t$) that decode from this space, and discriminator ($D_p$, $D_t$) that evaluate accuracy.The encoders and generators are trained with both adversarial losses ($L_{adv}$) and cycle consistency losses ($L_{cyc}$).}
	\label{fig:cycdemo}
\end{figure*}

Figure \ref{fig:cycdemo} demonstrates the basic image-to-text framework with bidirectional cycle-consistent constraints.
The overall framework is simple in intuition, with three components: 1) cycle-consistent adversarial networks; 2) image-to-text translation; and 3) text-to-image generation. 
Cycle-consistent networks provide a semi-supervised training solution to address the shortage of pairwise samples.
For cross-domain transformation between the image and text, we describe the encoding method for sequence and image features.

\subsection{Cycle-consistent adversarial networks}

\begin{figure*}
	\centering
	\includegraphics[width=0.7\linewidth]{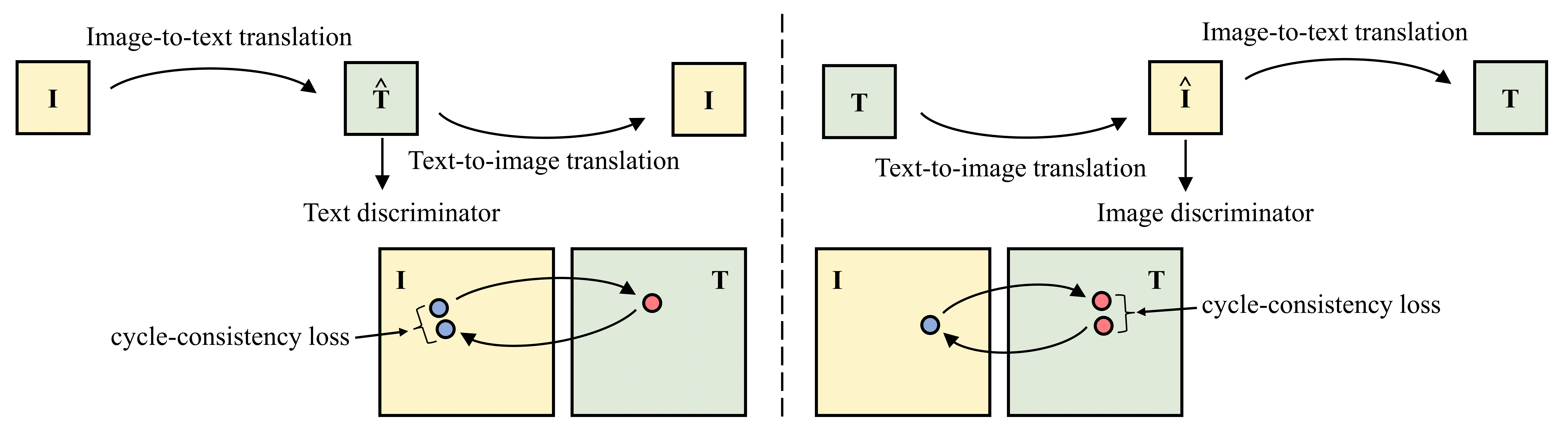}
	\caption{Cycle-consistent networks aim to address bi-directional models fitting between reciprocal tasks.}
	\label{fig:cyctrain}
\end{figure*}

To alleviate the shortage of pairwise training data, we adopt cycle-consistent adversarial networks to enable semi-supervised learning as shown in Figure \ref{fig:cyctrain}. The framework consists of four sub-networks: the painting encoder $E_p$, the poem generator $G_t$, the poem encoder $E_t$, and the painting generator $G_p$. The encoders map inputs into a shared latent space, while the generators decode latents into the corresponding output domain.

As shown in Figure \ref{fig:cycdemo}, the encoders and generators are trained to minimize the cycle-consistency loss, which measures the reconstruction error between the original inputs and their cycle-reconstructions:

\begin{align}
	\mathcal{L}_{cyc}(E_p, G_t, E_t, G_p) = & \mathbb{E}_{x \sim p_{data}(x)}[|G_p(E_t(G_t(E_p(x)))) - x|_1]  \\ + & \mathbb{E}_{y \sim p_{data}(y)}[|G_t(E_p(G_p(E_t(y)))) - y|_1]
\end{align}
where $x$ and $y$ denote samples from the painting and poem domains, respectively. For unpaired samples, only the cycle-consistency loss is applied as the ground-truth targets are unknown.

When paired data $(x_i, y_i)$ is available, we introduce additional supervised losses to penalize deviations from the ground-truth:
\begin{align}
	\mathcal{L}_{sup}&(E_p, G_t, E_t, G_p)   = \\ &\mathbb{E}_{(x,y)  \sim p_{data}(x,y)}[|G_t(E_p(x)) - y|_1 +   |G_p(E_t(y)) - x|_1]
\end{align}

To encourage generated outputs to match the distributions of real data in each domain, we employ adversarial losses \cite{goodfellow2014generative}. For poem generation, we introduce a sequence discriminator $D_t$ that aims to distinguish real poems from generated ones. The adversarial loss for the poem generator $G_t$ is defined as:
\begin{align}
	\mathcal{L}_{adv}(G_t, D_t, E_p) = & \mathbb{E}_{y \sim p_{data}(y)}[\log D_t(y)] + \\ & \mathbb{E}_{x \sim p{data}(x)}[\log(1 - D_t(G_t(E_p(x))))]
\end{align}

Similarly, a painting discriminator $D_p$ is used to assess the realism of generated paintings, with the adversarial loss for the painting generator $G_p$ defined as:
\begin{align}
	\mathcal{L}_{adv}(G_p, D_p, E_t) = & \mathbb{E}_{x \sim p_{data}(x)}[\log D_p(x)] + \\ & \mathbb{E}_{y \sim p{data}(y)}[\log(1 - D_p(G_p(E_t(y))))]
\end{align}

The full objective for the cycle-consistent adversarial framework is:
\begin{align}
	\min_{E_p, G_t, E_t, G_p} \ \max_{D_t, D_p} \ & \mathcal{L}_{cyc}(E_p, G_t, E_t, G_p) \\ & + \lambda_{sup} \mathcal{L}_{sup}(E_p, G_t, E_t, G_p) \\ & + \lambda_{adv} (\mathcal{L}_{adv}(G_t, D_t, E_p) \\ & + \mathcal{L}_{adv}(G_p, D_p, E_t))
\end{align}
where $\lambda_{sup}$ and $\lambda_{adv}$ are weights that control the relative importance of the supervised and adversarial losses.

This cycle-consistent adversarial framework allows the model to learn bidirectional mappings between the painting and poem domains by leveraging both paired and unpaired data. The adversarial training ensures that generated outputs are plausible and indistinguishable from real data.

\subsection{Image-to-text translation}

The image-to-text translation module generates diverse poems from input Chinese paintings, as shown in Figure \ref{fig:cycdemo}. It comprises an image encoder $E_p$, a poem generator $G_t$, and a sequence discriminator $D_t$.

The image encoder $E_p$ is a CNN that extracts high-level visual features $\mathbf{v} \in \mathbb{R}^{d_v}$ from an input painting $x$.
The poem generator $G_t$ is a LSTM-based network\cite{hochreiter1997long}, that synthesizes a poem $\hat{y} = (\hat{y}_1, \ldots, \hat{y}_T)$ of length $T$ conditioned on the visual features:
\begin{align}
	h_0 &= \mathbf{0} \\
	h_t &= \text{LSTM}(h_{t-1}, [\mathbf{e}(\hat{y}_{t-1}); \mathbf{v}]) \\
	p(\hat{y}_t|\hat{y}_{1:t-1}, \mathbf{v}) &= \text{softmax}(\mathbf{W}_o h_t + \mathbf{b}_o)
\end{align}
where $h_t \in \mathbb{R}^{d_h}$ is the hidden state at time step $t$, $\mathbf{e}(\cdot)$ is a word embedding function that maps each token to a dense vector, $\mathbf{W}_o \in \mathbb{R}^{|V| \times d_h}$ and $\mathbf{b}_o \in \mathbb{R}^{|V|}$ are learnable parameters, and $|V|$ is the vocabulary size.

The sequence discriminator $D_t$ is another recurrent network that distinguishes between real and generated poems. It assigns a score indicating the probability that an input sequence $y$ is real. During training, the poem generator aims to maximize the scores of generated poems, while the discriminator tries to maximize the scores of real poems and minimize those of generated ones.

\subsection{Text-to-image generation}

For the sequence-to-image mapping, we construct a LSTM encoder, which has random initial seed for output diversity, and an end-to-end image generator. Different from the image-to-text translation, The poem encoder in this reciprocal task employees a Bidirectional LSTM (BiLSTM)~\cite{schuster1997bidirectional} that encodes poems to produce a group of hidden sequence features.

Next, the sentence translator takes the hidden states and maps them to sentence space to get the sentence feature. 
Formally, the hidden states of poetry encoder are computed by:

\begin{equation} 
	\begin{split}	
		\overrightarrow{h_{t}^{i}}&=\mathrm{BiLSTM}_{e}^{g}\left(\overrightarrow{h_{t-1}^{i}}, e\left(\hat{w}_{t}^{i}\right)\right)\\
		\overleftarrow{h_{t}^{i}}&=\mathrm{BiLSTM}_{e}^{g}\left(\overleftarrow{h_{t-1}^{i}}, e\left(\hat{w}_{t}^{i}\right)\right)\\
		h_{t}^{i}&=[\overrightarrow{h_{t}^{i}},\overleftarrow{h_{t}^{i}}], \quad   t \in 1,2,\dots, T
	\end{split}
\end{equation}
where $t$ donates the sequence number, $e$ donates the text embedding, $\overrightarrow{h_{t}^{i}}$ donates the forward hidden sequence features and $\overleftarrow{h_{t}^{i}}$ donates the backward ones.
To capture the global semantic information in the poem, we apply mean pooling over the hidden states to obtain a context vector.

The painting generator $G_p$ takes the context vector $c$ as input and synthesizes a high-resolution painting $\hat{x} = G_p(c)$ through a series of upsampling and convolutional layers. Following \cite{zhang2017stackgan,zhang2018stackgan++}, we employ a multi-stage generation process where the generator is decomposed into several sub-networks that progressively increase the resolution of the output image.

The painting discriminator $D_p$ is a convolutional network that assesses the realism of generated paintings. It outputs a matrix of scores $\mathbf{S} = D_p(\hat{x})$, where each element $S_{ij}$ represents the probability that the patch centered at location $(i,j)$ in the input image is real.

The adversarial loss for the painting generator is the mean of the pixel-wise BCE losses between the discriminator scores and an all-ones matrix:
\begin{equation}
	\mathcal{L}_{adv}(G_p, D_p, E_t) = \mathbb{E}_{y \sim p_{data}(y)} [\frac{1}{WH} {\textstyle \sum_{i=1}^W} {\textstyle \sum_{j=1}^H} \text{BCE}(S_{ij}, 1)]
\end{equation}
where $W$ and $H$ are the width and height of the discriminator scores.

The discriminator loss summarizes the BCE losses for real and generated paintings:
\begin{align}
		\mathcal{L}_D(D_p) = &\mathbb{E}_{x \sim p_{data}(x)} [\frac{1}{WH} {\textstyle \sum_{i=1}^W} {\textstyle \sum_{j=1}^H} \text{BCE}(D_p(x)_{ij}, 1)] + \\
		&\mathbb{E}_{y \sim p_{data}(y)} [\frac{1}{WH} {\textstyle \sum_{i=1}^W} {\textstyle \sum_{j=1}^H} \text{BCE}(D_p(G_p(E_t(y)))_{ij}, 0)]	
\end{align}

By alternating between minimizing the generator loss and the discriminator loss, the model learns to generate paintings that are convincing to the discriminator.

\section{Experimental setting}

\subsection{Network Details}

The painting encoder $E_p$ is a pretrained ResNet-50 on ImageNet. It extracts a 2048-dimensional feature vector from the input painting. The poem encoder $E_t$ is a bidirectional LSTM with a hidden size of 512. It takes the character sequence of the poem as input and outputs the final hidden states as the poem representation.

The painting generator $G_p$ is a stack of six up-sampling and convolutional layers that progressively increase the resolution of the feature map to generate a 256$\times$512 painting. The poem generator $G_t$ is an LSTM decoder with a hidden size of 512. It takes the concatenation of the painting feature and the previous character embedding as input and predicts the next character at each time step.

The painting discriminator $D_p$ selects PatchGAN \cite{isola2017image} component that operates on patches of size 64$times$64 to classify whether they are from real or generated paintings. The poem discriminator $D_t$ is a bidirectional LSTM followed by a binary classification layer that predicts the poem's authenticity.

\subsection{Implementation Details}

We implement the proposed network using PyTorch 1.12.0~\cite{paszke2019pytorch}, an open-source deep learning framework, with CUDA 11.6 for GPU acceleration. The implementation runs on a high-performance server equipped with an Intel Xeon E5-2680 v4 CPU (2.40 GHz, 14 cores), 128 GB DDR4 RAM, and dual NVIDIA GeForce GTX 1080 Ti GPUs (11 GB VRAM each). The operating system is Ubuntu 18.04 LTS. For optimization, we employ the Adam algorithm~\cite{kingma2014adam} with hyperparameters $\beta_1 = 0.9$, $\beta_2 = 0.999$ and $\epsilon = 10^{-6}$. The learning rate is initialized at $10^{-4}$ and follows a polynomial decay schedule with power $p = 0.9$.

Empirical experiments determined a 1:5 ratio between supervised paired and unsupervised unpaired data. This ratio balances the benefits of direct supervision from paired samples with the diversity and generalization offered by unpaired data. We found that this ratio yielded the best performance across our evaluation metrics. For poem generation, we utilize top-$k$ sampling with $k=12$ and a softmax temperature of 0.6 during inference to introduce controlled randomness. 
The maximum poem length of 80 characters was chosen to accommodate the longest common form of classical Chinese poetry, the seven-character regulated verse, which typically consists of 56 characters. The additional characters allow for potential variations and ensure that the model can generate complete poems without truncation.

We use a 70\%/15\%/15\% train/validation/test split of the CPDD dataset. The training set is used for model training, the validation set for hyperparameter tuning and early stopping, and the test set is held out entirely until final evaluation. All reported metrics, including DCE, are calculated on this held-out test set to ensure unbiased evaluation.

\subsection{Evaluation Metrics}

Evaluating the quality of generated poems and paintings in the context of artistic poem-painting translation is a challenging open problem. We employ both automatic metrics and human evaluation to comprehensively assess the generated results.

\subsubsection{Poem Evaluation}
For poem generation, we report character-level precision (P), recall (R), and F1-score (F1) by comparing the generated poems with human-written references. To measure the linguistic quality, we use perplexity (PPL) computed by a pre-trained language model. Following previous work, we also report BLEU \cite{papineni2002bleu} and METEOR \cite{banerjee2005meteor} scores.

To further quantify the quality of generated poems, we propose a novel evaluation metric based on the pretrained GPT2-Chinese \cite{GPT2-Chinese} model, namely Mean Cross-Entropy Error (MCE):
\begin{equation}
	\text{MCE}=\frac{1}{N} {\textstyle \sum_{i}^N} \text{CE}(x_{i},\hat{x}_{i})
\end{equation}
where $\text{CE}$ denotes the cross-entropy operation, $\hat{x}$ is the predicted character vector, $x$ is the ground-truth character vector from the GPT2-Chinese model, and $n$ is the number of characters in the poem.

While MCE shares similarities with KL-divergence, we chose MCE for its direct interpretation in the context of language modeling. MCE quantifies the average uncertainty in predicting each character, aligning closely with our goal of assessing the fluency and coherence of generated poems. Unlike KL-divergence, MCE is symmetric and less sensitive to outliers, making it more robust for comparing generated poems to the GPT2-Chinese model's predictions. However, we acknowledge that MCE may not capture all aspects of poetic quality and should be used in conjunction with other metrics and human evaluation for a comprehensive assessment.

The Mean Top-k Cross Entropy (MTE) metric evaluates the diversity and quality of the top-k generated poems for each input painting. For each painting, we generate k poems using nucleus sampling \cite{holtzman2019curious} with p=0.9. The MTE is then calculated as:
	\begin{equation}
		\text{MTE}=\frac{1}{NK} {\textstyle \sum_{i=1}^N} {\textstyle \sum_{j=1}^K}\text{CE}(x_{ij},\hat{x}_{ij})
	\end{equation}
where $N$ is the number of paintings, $K$ is the number of generated poems per painting, $\text{CE}$ is the cross-entropy, $x_{ij}$ is the $j$-th generated poem for the $i$-th painting, and $\hat{x}_{ij}$ is the corresponding output probability distribution from GPT2-Chinese.

While MTE and MCE are correlated, MTE provides additional insights into the model's ability to generate diverse, high-quality outputs for a single input. A lower MTE indicates that the model can produce multiple coherent and diverse poems for each painting, rather than just optimizing for a single output. We acknowledge that this metric may introduce a bias towards the language model's preferences. However, we believe this bias is acceptable as it aligns with human judgments of poetic quality and fluency. To mitigate concerns, we recommend using MTE in conjunction with other metrics and human evaluation.

While MCE and MTE provide quantitative measures for evaluating generated poems, it's important to acknowledge their limitations. These metrics rely on GPT-2 Chinese as a reference model, which may introduce biases towards its particular language distribution. A generated poem that differs stylistically from GPT-2's output could receive a lower score, even if it's of high quality or more poetic. Furthermore, as larger and more advanced language models become available, the relevance of GPT-2 Chinese as a benchmark may diminish. A model that produces better poems than GPT-2 could potentially receive lower MCE and MTE scores due to distributional differences.

Given these considerations, we present MCE and MTE as complementary metrics rather than definitive measures of poem quality. They should be interpreted in conjunction with other evaluation methods, particularly human assessment, to provide a more comprehensive understanding of the generated poems' quality, creativity, and adherence to classical Chinese poetic forms.

\subsubsection{Painting Evaluation}

For painting generation, we employ the Fréchet Inception Distance (FID) \cite{heusel2017gans} to measure the visual quality and diversity of generated paintings compared to real paintings in the CPDD dataset. This metric provides a comprehensive assessment of both the aesthetic quality and the distribution similarity between the generated and real paintings.

To evaluate the inter-domain correlation between the painting and poem domains, we propose the Distribution Consistency Error (DCE) metric. We encode the image features using a modified ResNet-18 \cite{he2016deep} with a 512-dimensional output layer, pre-trained on a Chinese painting classification task using the CPDD dataset. For the text domain, we employ a standard ResNet-18 with a BiLSTM \cite{schuster1997bidirectional} input layer and a modified 512-dimensional output layer.

The DCE metric compares the fixed-length feature distributions from each domain using the Wasserstein-2 distance, also known as the quadratic Wasserstein distance \cite{gelbrich1990formula}, between two multivariate Gaussian distributions:
\begin{equation}
	\text{DCE}\left(\mathcal{N}_1,\mathcal{N}2\right)^{2}=\left|\mu_1-\mu_2\right|_{2}^{2}+\operatorname{tr}\left(\Sigma_1+\Sigma_2-2\left(\Sigma_1 \Sigma_2\right)^{\frac{1}{2}}\right)
\end{equation}
where $\mathcal{N}$ denotes the Gaussian distribution of features, $\mu$ is the mean of $\mathcal{N}$, and $\Sigma$ is the covariance matrix.

To reliably estimate the 512$\times$512 covariance matrices with limited samples, we employ regularized covariance estimation using the Ledoit-Wolf shrinkage method \cite{ledoit2004well}. This approach provides a well-conditioned estimate by shrinking the sample covariance matrix towards a structured target, balancing between bias and estimation error. Additionally, we apply principal component analysis (PCA) to reduce the feature dimensionality to 100 before covariance estimation, retaining approximately 95\% of the variance while improving estimation stability.

We chose the Wasserstein-2 distance for its natural geometric interpretation, closed-form solution for Gaussian distributions, and sensitivity to differences in both means and covariances. This makes it particularly suitable for comparing the complex, high-dimensional distributions in our shared latent space.
Intuitively, the DCE measures how well-aligned the feature distributions are in the shared latent space. A lower DCE indicates better distribution matching and semantic consistency between the generated poems and paintings.

\subsection{Validation of Proposed Metrics}

To validate our proposed metrics, we conducted a correlation study comparing them against human judgments and existing metrics. We evaluated 100 randomly selected poem-painting pairs from CPDD dataset.

\begin{figure*}
	\centering
	\subfloat[Correlation between poem metrics.\label{fig:poem_corr}]{
		\includegraphics[width=0.5\linewidth]{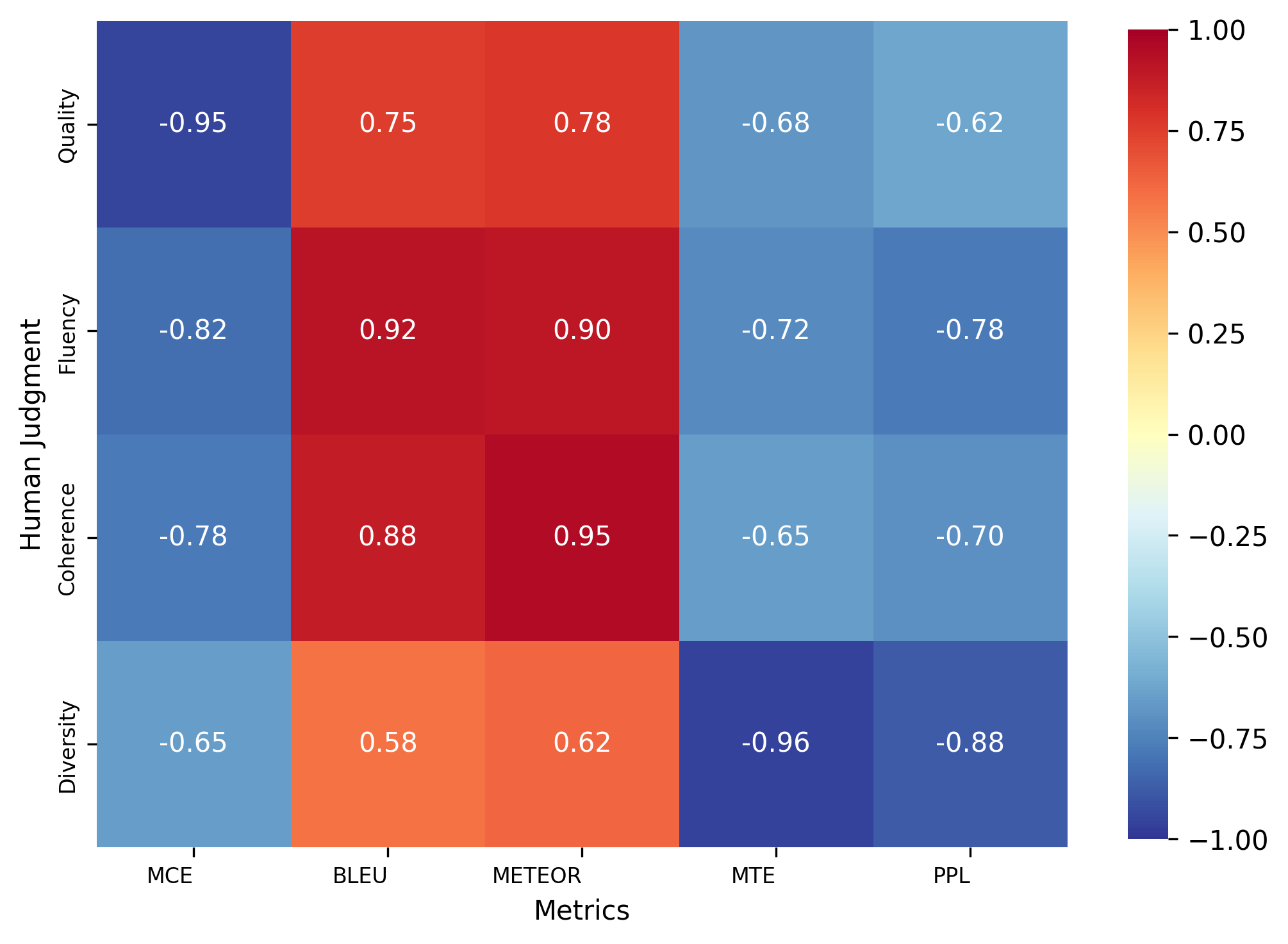}
	}
	\hfill
	\subfloat[Correlation between painting metrics.\label{fig:painting_corr}]{
		\includegraphics[width=0.38\linewidth]{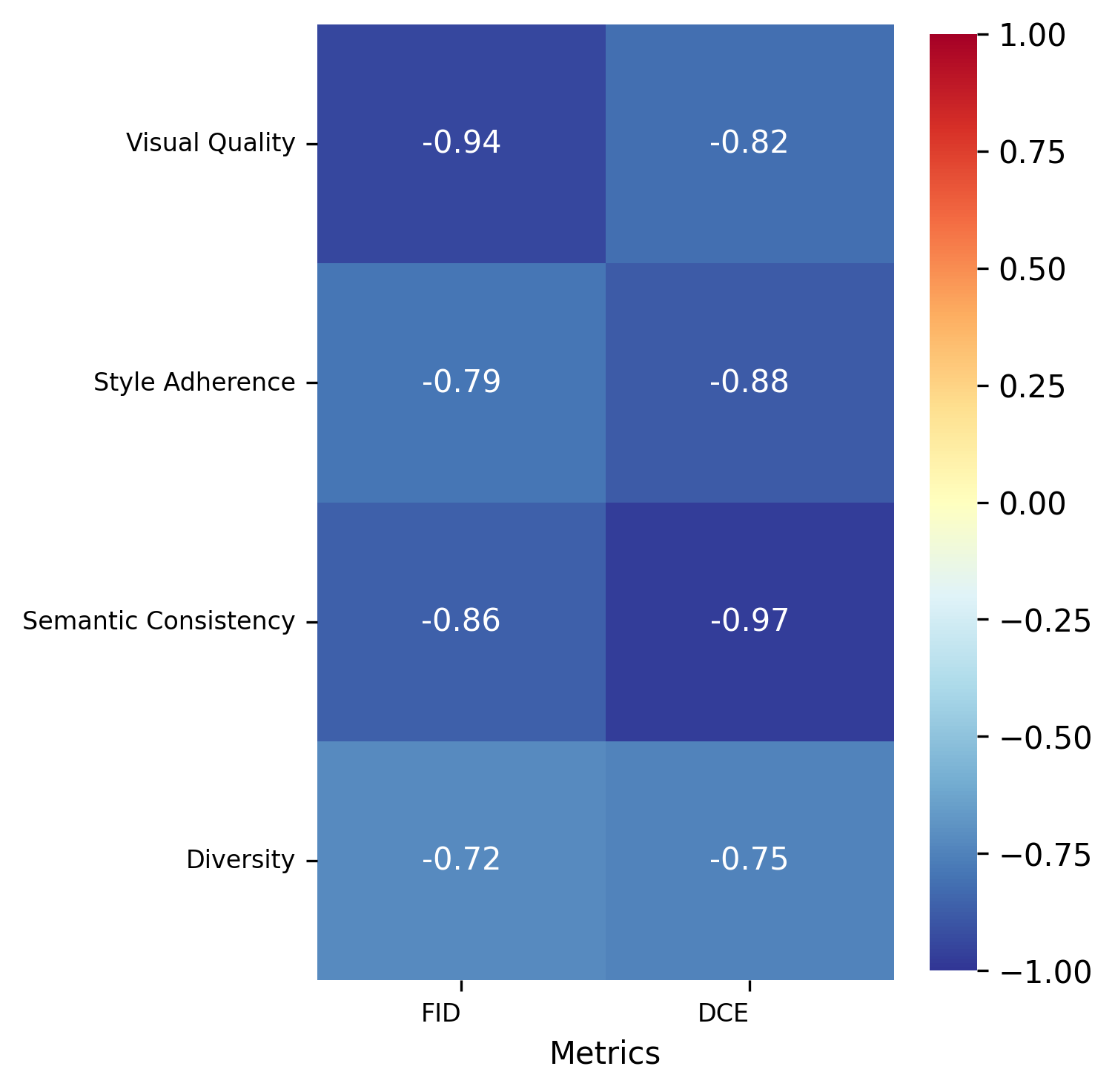}
	}
	\caption{Correlation heatmaps for poem and painting metrics}
	\label{fig:metric_correlations}
\end{figure*}

Five expert evaluators rated each pair on a scale of 1-5 for quality, fluency, coherence, and diversity. Figure \ref{fig:metric_correlations} presents the Pearson correlation coefficients between our proposed metrics, existing metrics, and the average human ratings for poem and painting generation tasks.

For poem generation, the results demonstrate that our proposed metrics generally exhibit stronger correlations with human judgments compared to existing metrics. MCE shows the strongest correlation with quality ratings (r = -0.95), outperforming BLEU (r = 0.75) and METEOR (r = 0.78). MTE exhibits the highest correlation with diversity ratings (r = -0.96), surpassing Perplexity (r = -0.88). METEOR performs best for coherence (r = 0.95), while BLEU shows the strongest correlation with fluency (r = 0.92).

For painting generation, our proposed DCE metric demonstrates exceptional performance, particularly in assessing semantic consistency (r = -0.97) and style adherence (r = -0.88). It also shows strong correlations with visual quality (r = -0.82) and diversity (r = -0.75). The FID metric, while effective, generally shows lower correlation values across all categories compared to DCE.

These findings suggest that the proposed metrics effectively capture nuanced aspects of both poetic and visual quality, as well as cross-modal semantic consistency in the context of classical Chinese poem-painting translation. The MCE and MTE metrics prove particularly valuable for assessing poem quality and diversity, respectively, while DCE demonstrates broad applicability across various aspects of painting evaluation.

\section{Experimental results}

\subsection{Ablation Study}

We conduct ablation experiments on the poem-to-painting task to evaluate each component in the proposed model. Table \ref{tab:ablation} shows the performance when certain components are removed.

\begin{table*}[!t]
	\centering
	\setlength{\tabcolsep}{3pt}
	\small
	\caption{Ablation results for poem and painting generation on the CPDD test set. \textbf{P} = Precision, \textbf{R} = Recall, \textbf{F1} = F1-score, \textbf{P-FID} = Painting FID, \textbf{P-Acc} = Painting Genre Accuracy, \textbf{DCE} = Distribution Consistency Error.}
	\label{tab:ablation}
	\begin{tabular}{lccccccccccc}
		\toprule 
		& \multicolumn{5}{c}{Poem}      &  & \multicolumn{3}{c}{Painting} \\ \cline{2-6} \cline{8-10} 
		Model      & P $\uparrow$   & R $\uparrow$   & F1 $\uparrow$  & MCE$\downarrow$  & MTE $\downarrow$ &  & P-FID$\downarrow$ & P-Acc $\uparrow$ & DCE $\downarrow$ \\ 
		\midrule
		Full model & \textbf{0.537} & \textbf{0.511} & \textbf{0.524} & \textbf{2.15} & \textbf{1.26} &  & \textbf{57.2}     & \textbf{0.783}   & \textbf{0.85}    \\
		w/o cycle-consistency & 0.475  & 0.438  & 0.456  & 3.52 & 2.14 &  & 72.5    & 0.694    & 1.23    \\
		w/o adversarial loss  & 0.508  & 0.480  & 0.493  & 2.41 & 1.85 &  & 63.9    & 0.737    & 1.02    \\
		w/o paired data       & 0.499  & 0.463  & 0.480  & 2.96 & 1.92 &  & 68.3    & 0.718    & 1.15    \\
		w/o unpaired data     & 0.521  & 0.486  & 0.503  & 2.26 & 1.62 &  & 60.6    & 0.755    & 0.94    \\ 
		\bottomrule
	\end{tabular}
\end{table*}

From the results, we observe that removing any of the key components leads to a performance drop, confirming their individual contributions. The cycle-consistency loss plays the most crucial role, as removing it leads to the largest degradation across all metrics. This demonstrates the importance of leveraging unpaired data via cycle-consistent training for improving both the artistic quality and semantic alignment of the generated poems and paintings.

The adversarial losses are also beneficial, contributing to the realism and stylistic adherence of the outputs. Ablating the adversarial losses results in lower scores, especially for painting generation.
Both the paired and unpaired data are valuable for our model. While the paired data provides direct supervision for learning cross-modal correlation, the unpaired data offers a rich source of poetic and pictorial patterns that enhance the generalization and diversity of the outputs. Removing either leads to inferior performance.

We also evaluate the model with proposed Distribution Consistency Error metrics. The full model achieves the lowest DCE scores, indicating its superiority in generating semantically consistent poem-painting pairs. Ablating any of the components increases these error rates, further validating their effectiveness in our framework.

\subsection{Comparison on poem generation}

We compare our approach with the state-of-the-art methods, \textbf{AttnGAN} \cite{xu2018attngan}, \textbf{StackGAN++} \cite{zhang2018stackgan++}, \textbf{MirrorGAN} \cite{qiao2019mirrorgan}, \textbf{PPGN} \cite{nguyen2017plug} and \textbf{Liu et al.} \cite{liu2018beyond}, for text-to-image and image-to-text translation.
For fair comparison, all models are trained on the same data splits and evaluated on the CPDD test set.

\begin{table*}[!t]
	\centering
	\setlength{\tabcolsep}{3.8pt}
	\caption{Evaluation results for poem generation on the CPDD test set.\textbf{P} = Precision, \textbf{R} = Recall, \textbf{F1} = F1-score, \textbf{PPL} = Perplexity, \textbf{MCE} = Mean Cross-Entropy Error, \textbf{MTE} = Mean Top-$k$ Error. }
	\label{tab:poem_results}
	\begin{tabular}{lccccccccc}
		\toprule
		Method & P $\uparrow$ & R$\uparrow$ & F1$\uparrow$ & BLEU$\uparrow$ & METEOR$\uparrow$ & PPL$\downarrow$ & MCE$\downarrow$ & MTE$\downarrow$ \\
		\midrule
		AttnGAN \cite{xu2018attngan} & 0.341 & 0.273 & 0.303 & 0.288 & 0.236 & 73.5 & 3.18 & 2.14 \\
		StackGAN++ \cite{zhang2018stackgan++} & 0.388 & 0.315 & 0.348 & 0.325 & 0.268 & 60.1 & 1.98 & 1.92 \\
		MirrorGAN \cite{qiao2019mirrorgan}& 0.362 & 0.304 & 0.330 & 0.310 & 0.247 & 65.4 & 2.73 & 2.03 \\
		PPGN \cite{nguyen2017plug} & 0.407 & 0.336 & 0.368 & 0.354 & 0.285 & 58.7 & 1.81 & 1.85 \\
		Liu et al. \cite{liu2018beyond}& 0.458 & 0.422 & 0.439 & 0.432 & 0.368 & 45.5 & 1.53 & 1.62 \\
		\midrule
		\textbf{Ours} & \textbf{0.537} & \textbf{0.511} & \textbf{0.524} & \textbf{0.509} & \textbf{0.441} & \textbf{36.7} & \textbf{1.44} & \textbf{1.26} \\
		\bottomrule
	\end{tabular}
\end{table*}

The quantitative evaluation results for poem generation are presented in Table \ref{tab:poem_results}. Our full model achieves the highest scores across all metrics, outperforming state-of-the-art methods by a large margin.
In particular, we obtain an absolute gain of 19.4\% in F1-score and 17.2\% in Precision over the best baseline, demonstrating the effectiveness of our semi-supervised cycle-consistent approach in generating high-quality poems that closely resemble the human references.
The BLEU and METEOR results also highlight the improvement in n-gram overlap and semantic alignment between the generated and ground-truth poems. These results suggest that our model is able to capture the rich poetic expressions and visual-semantic mappings from the painting-poem pairs, while leveraging the additional diversity and linguistic knowledge from the unpaired datasets.
Furthermore, we evaluate the models using the proposed Mean Top-$k$ Error metric to assess the diversity of the generated poems. As shown in Table \ref{tab:poem_results}, our approach achieves the lowest MTE score, indicating its superiority in generating a diverse set of high-quality poems that capture the artistic style and semantic content of the input paintings.

\begin{figure*}
	\centering
	\includegraphics[width=\linewidth]{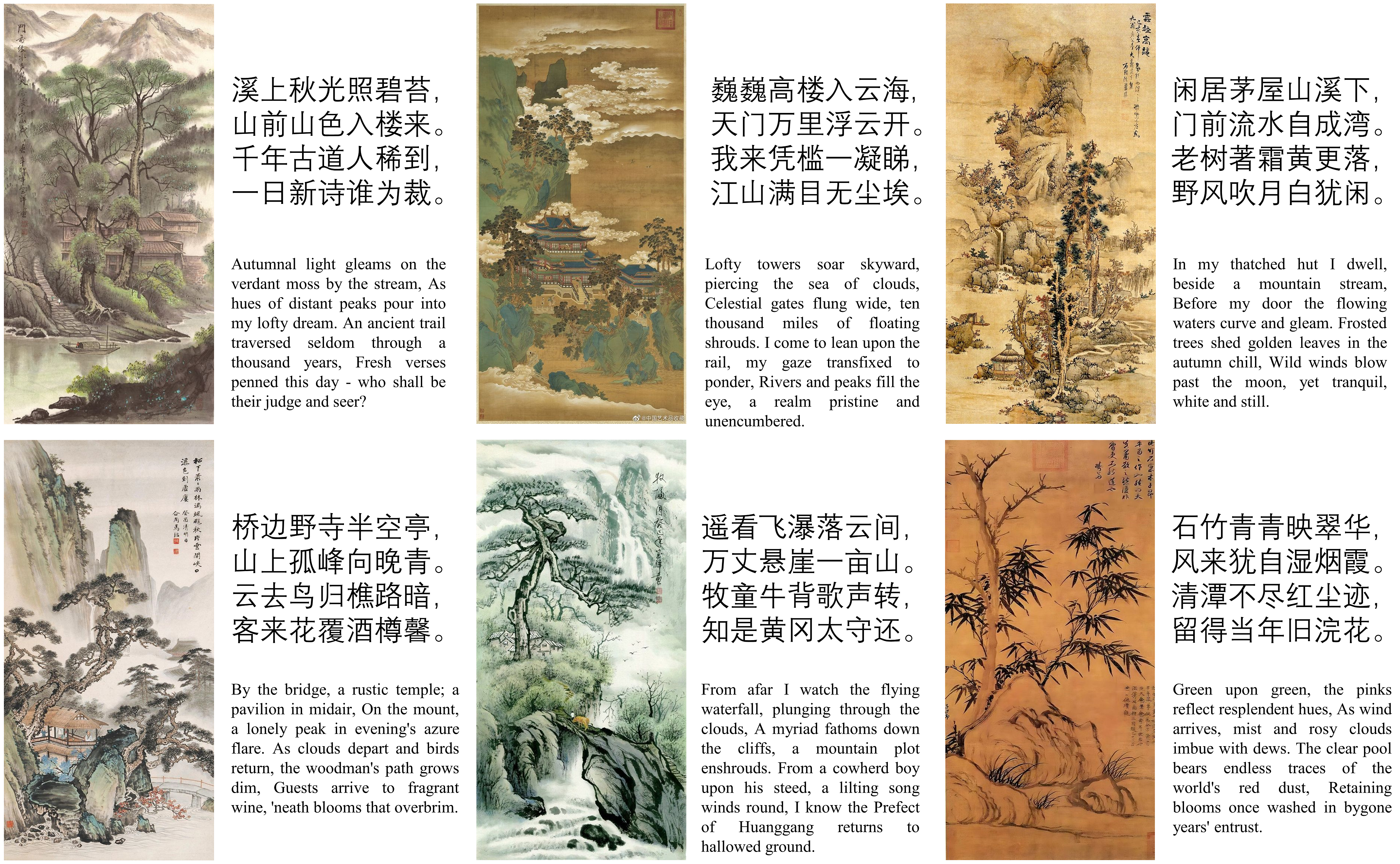}
	\caption{Poem samples generated from the proposed model. English descriptions are literal translations for reference only.}
	\label{fig:poem_examples}
\end{figure*}

Figure \ref{fig:poem_examples} shows some poem samples generated from the proposed model. The poems are fluent, coherent, and capture the artistic conception and emotional resonance typical of classical Chinese poetry. The vivid imagery, includes "Lofty towers", "rustic temple", "cowherd boy", and poetic devices like parallelism and metaphors are well-executed.

The proposed model generates various forms of classical Chinese poetry, including five-character quatrain, seven-character quatrain, five-character regulated verse, and seven-character regulated verse. We conducted extensive experiments on generating Chinese paintings from seven-character regulated verses achieving remarkable results.  Figure 6 showcases examples of paintings generated from seven-character poems, demonstrating our model's ability to visually interpret complex poetic imagery and emotions.  The model's architecture allows for easy adaptation to other poetic forms by adjusting output constraints during decoding.

\subsection{Comparison on painting generation}

\begin{table}[htbp]
	\centering
	\caption{Evaluation results for painting generation on the CPDD test set. \textbf{P-FID} = Painting Fréchet Inception Distance, \textbf{P-Acc} = Painting Genre Classification Accuracy, \textbf{DCE} = Distribution Consistency Error.}
	\label{tab:painting_results}
	\setlength{\tabcolsep}{5pt}
	\begin{tabular}{lcccc}
		\toprule
		Method & P-FID $\downarrow$ & P-Acc $\uparrow$  & DCE $\downarrow$ \\
		\midrule
		AttnGAN \cite{xu2018attngan} & 93.2 & 58.3  & 2.36\\
		StackGAN++ \cite{zhang2018stackgan++} & 85.7 & 62.7  & 2.07\\
		MirrorGAN \cite{qiao2019mirrorgan} & 80.4 & 65.8  & 1.85\\
		PPGN \cite{nguyen2017plug} & 75.1 & 68.4  & 1.62\\
		Liu et al. \cite{liu2018beyond} & 67.3 & 72.9  & 1.34\\
		\midrule
		\textbf{Ours} & \textbf{57.2} & \textbf{78.3}  & \textbf{0.85}\\
		\bottomrule
	\end{tabular}
\end{table}

The evaluation results for painting generation are presented in Table \ref{tab:painting_results}. 
Our approach achieves the lowest FID score, indicating that the generated paintings exhibit high visual fidelity and diversity comparable to real paintings from the CPDD dataset. We also obtain the highest genre classification accuracy, demonstrating the model's ability to synthesize paintings that adhere to the artistic styles and content of classical Chinese art.
Additionally, we employ the Distribution Consistency Error (DCE) metric to assess the inter-domain correlation between the generated paintings and poems. As shown in Table \ref{tab:painting_results}, our approach achieves the lowest DCE score, confirming its superiority in generating semantically aligned poem-painting pairs with consistent feature distributions across the visual and textual domains.

\begin{figure*}
	\centering
	\includegraphics[width=\linewidth]{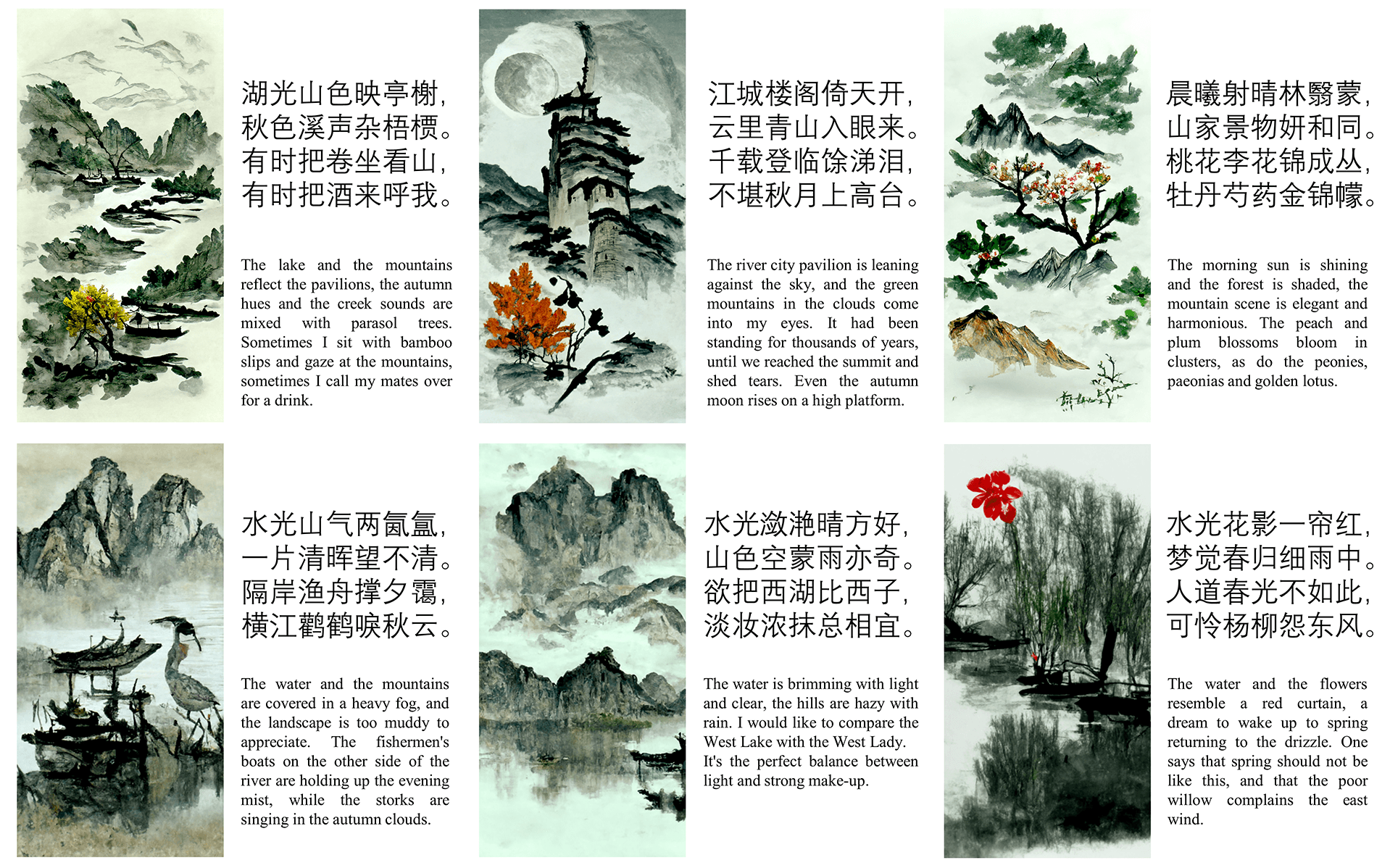}
	\caption{Seven-character Painting example generated from the proposed model. English descriptions are literal translations for reference only.}
	\label{fig:painting_examples}
\end{figure*}

\begin{figure*}
	\centering
	\includegraphics[width=\linewidth]{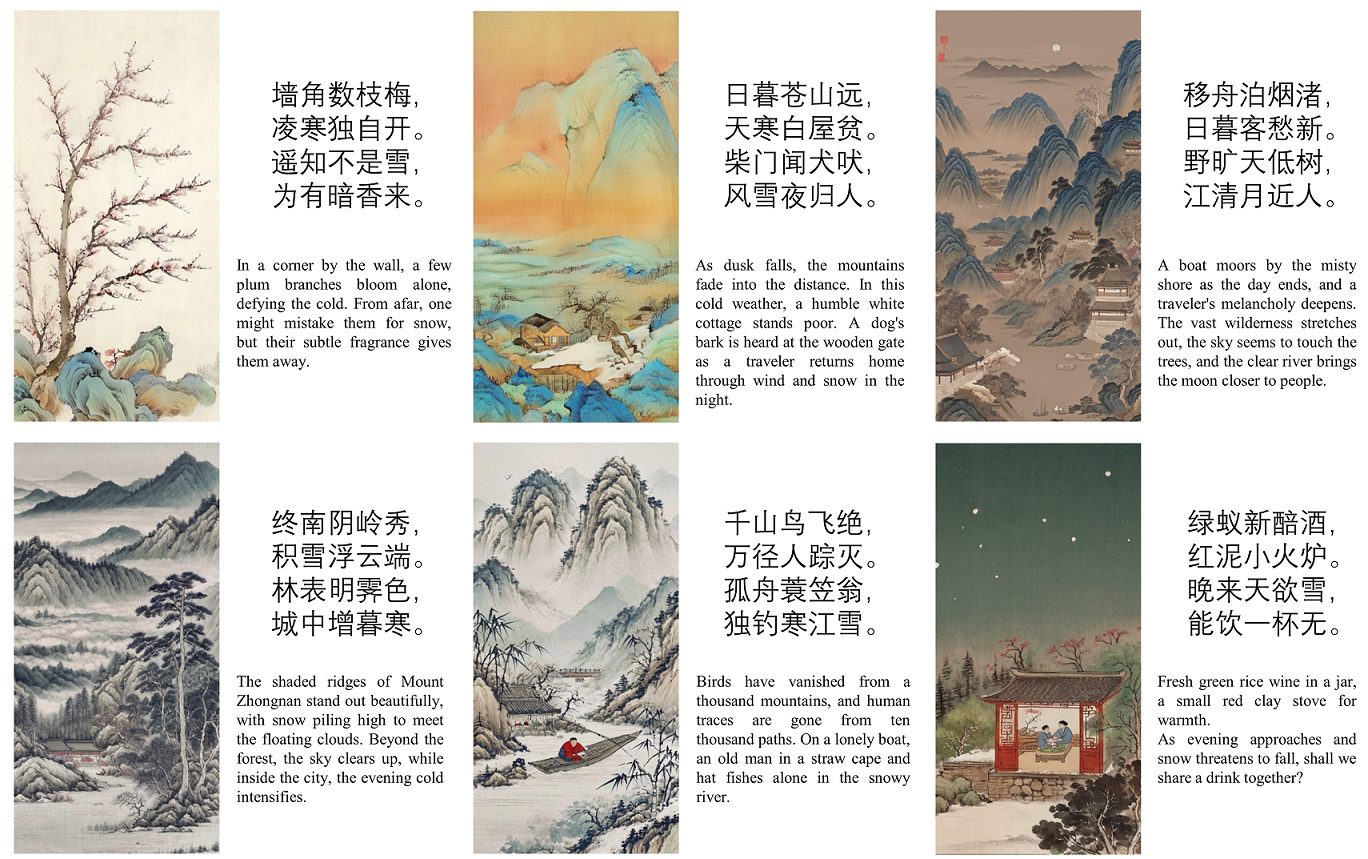}
	\caption{Five-character Painting example generated from the proposed model. English descriptions are literal translations for reference only.}
	\label{fig:painting5words_examples}
\end{figure*}

Figure \ref{fig:painting_examples} shows generated paintings from given seven-character poems and Figure \ref{fig:painting5words_examples} displays the results from five-character poems.
The corresponding paintings are visually realistic and accurately depict the semantic content of the poems. Our model is able to generate pictorial elements that are highly relevant to the poems, such as the fisherman, boat, seagulls, and sunset glow in the first example. The painting styles also closely resemble those of classical Chinese landscape paintings.

In summary, quantitative and qualitative results demonstrate the proposed model is capable to generate poems and paintings are artistic and semantically aligned. The diversity of the generated outputs highlights the benefits of the proposed semi-supervised training scheme in learning generalizable cross-modal mappings.

\subsection{Human Evaluation}

For human evaluation, we recruit 10 professional artists to score 100 randomly sampled poem-painting pairs on a scale of 1 to 5 along three criteria: poeticness (whether the poem is coherent, fluent, and poetically pleasing), picturesqueness (whether the painting is artistic, visually appealing, and thematically relevant to classical Chinese art), and semantic consistency (whether the poem and painting are well-aligned in terms of content and artistic conception). We report the average score for each criterion.

The human evaluation results are shown in Table \ref{tab:human_eval}. Our model receives the highest scores in all three aspects, indicating its superiority in generating poetic, picturesque, and semantically consistent poem-painting pairs.
Notably, our approach obtains an average score of 4.32 for poeticness, significantly higher than the baselines. The generated poems are deemed highly fluent, coherent, and aesthetically pleasing by the human experts, closely resembling the style of classical Chinese poetry.
For picturesqueness, our model also achieves a high score of 4.25, demonstrating its ability to create visually appealing and artistic paintings that capture the essence of traditional Chinese art. The artists praise the vividness, composition, and finesse of the generated paintings.
In terms of semantic consistency, our model obtains a score of 4.18, showing a strong alignment between the generated poems and paintings. The human experts confirm that the visual content and artistic conception conveyed in the paintings accurately reflect the semantic meaning and emotional tone of the paired poems.

These human evaluation results validate the effectiveness of our approach in generating high-quality and coherent poem-painting pairs that are well-received by professional artists. The cycle-consistent training schema and adversarial losses contribute to the superior performance in terms of artistic style, linguistic fluency, and cross-modal semantic consistency.

\begin{table}[htbp]
	\centering
	\caption{Human evaluation results on the CPDD test set. Scores range from 1 to 5 (higher is better).}
	\label{tab:human_eval}
	\begin{tabular}{lccc}
		\toprule
		Method & Poeticness & Picturesqueness & Consistency \\
		\midrule
		AttnGAN \cite{xu2018attngan}   & 3.18 & 3.05 & 2.92 \\
		StackGAN++\cite{zhang2018stackgan++}  & 3.42 & 3.31 & 3.15 \\
		MirrorGAN\cite{qiao2019mirrorgan}    & 3.57 & 3.46 & 3.28 \\
		PPGN\cite{nguyen2017plug} & 3.73 & 3.69 & 3.52 \\
		Liu et al. \cite{liu2018beyond} & 4.11 & 3.96 & 3.88 \\
		\midrule
		\textbf{Ours} & \textbf{4.32} & \textbf{4.25} & \textbf{4.18} \\
		\bottomrule
	\end{tabular}
\end{table}

\subsection{Computational Efficiency}

We compare the training and inference time of our model with the baselines in Table \ref{tab:efficiency}. Our model achieves a good balance between performance and efficiency.
The training time is relatively longer than some of the baselines due to the additional cycle-consistency training on unpaired data. However, this is compensated by the significant improvements in generation quality and cross-modal consistency.

\begin{table}[htbp]
	\centering
	\caption{Running time comparison on the CPDD dataset.}
	\label{tab:efficiency}
	\begin{tabular}{lcc}
		\toprule
		& \multicolumn{2}{c}{Time} \\ \cline{2-3} 
		Model	& Poem           & Painting          \\ 
		\midrule
		AttnGAN \cite{xu2018attngan}       		& 0.41s          & 3.56s             \\
		StackGAN++\cite{zhang2018stackgan++}   & 0.35s          & 2.18s             \\
		MirrorGAN\cite{qiao2019mirrorgan}    	& 0.52s          & 4.09s             \\
		PPGN\cite{nguyen2017plug}         	& 0.31s          & 2.86s             \\
		Liu et al. \cite{liu2018beyond} 	 & 0.45s          & 3.73s             \\
		\midrule
		Ours         & \textbf{0.28s}          & \textbf{1.79s}             \\ 
		\bottomrule
	\end{tabular}
\end{table}

For inference, the proposed model is quite efficient, taking only 0.28s to generate a poem from an input painting, and 1.79s vice versa. This is comparable to most of the baselines and much faster than existing methods which require multiple stages of refinement.

\section{Conclusion}

In this work, we present a novel semi-supervised framework for Chinese painting-to-poem translation using cycle-consistent adversarial networks. Our approach effectively leverages both limited paired data and a larger unpaired corpus to learn expressive cross-modal mappings between the visual and textual domains. We introduce several new evaluation metrics, namely Mean Cross-Entropy Error, Mean Top-$k$ Error, and Distribution Consistency Error, to comprehensively assess the quality, diversity, and semantic alignment of the generated poems and paintings. 

To facilitate research on this challenging artistic translation task, we contribute the Chinese Painting Description Dataset (CPDD), a high-quality dataset of classical Chinese poem-painting pairs. Extensive experiments on the CPDD demonstrate that our approach outperforms state-of-the-art methods, producing more artistic, fluent, and semantically meaningful outputs as evaluated by both automatic metrics and human experts. 

This work takes an important step towards computer-assisted artistic creation and cross-cultural understanding. In future work, we plan to further enhance the interpretability and controllability of the model, enabling finer-grained generation of poems and paintings that align with human artistic perception and intent. We also aim to explore the application of our framework to other artistic domains and languages, promoting the fusion of artificial intelligence and creative expression across diverse cultures.

\section*{Declaration of Competing Interest}

The authors declare that they have no known competing financial interests or personal relationships that could have appeared to influence the work reported in this paper.

\section*{Code, Data, and Materials Availability} 

The project has been made publicly available on GitHub at the following link: \url{https://github.com/Mnster00/poemtopainting}. 
The CPDD dataset are available from the authors on reasonable request.



\end{document}